# Task-Invariant Learning of Continuous Joint Kinematics during Steady-State and Transient Ambulation Using Ultrasound Sensing

M. Hassan Jahanandish, Kaitlin G. Rabe, *Student Member, IEEE,* Abhishek Srinivas, Nicholas P. Fey, *Member, IEEE*, Kenneth Hoyt*, *Senior Member, IEEE*

*Abstract*—Natural control of limb motion is continuous and progressively adaptive to individual intent. While intuitive interfaces have the potential to rely on the neuromuscular input by the user for continuous adaptation, continuous volitional control of assistive devices that can generalize across various tasks has not been addressed. In this study, we propose a method to use spatiotemporal ultrasound features of the rectus femoris and vastus intermedius muscles of able-bodied individuals for task-invariant learning of continuous knee kinematics during steady-state and transient ambulation. The task-invariant learning paradigm was statistically evaluated against a task-specific paradigm for the steady-state (1) level-walk, (2) incline, (3) decline, (4) stair ascent, and (5) stair descent ambulation tasks. The transitions between steady-state stair ambulation and level-ground walking were also investigated. It was observed that the continuous knee kinematics can be learned using a task-invariant learning paradigm with statistically comparable accuracy to a task-specific paradigm. Statistical analysis further revealed that incorporating the temporal ultrasound features significantly improves the accuracy of continuous estimations ($p < 0.05$). The average root mean square errors (RMSEs) of knee angle and angular velocity estimation were 7.06° and 53.1°/sec, respectively, for the task-invariant learning compared to 6.00° and 51.8°/sec for the task-specific models. High accuracy of continuous task-invariant paradigms overcome the barrier of task-specific control schemes and motivate the implementation of direct volitional control of lower-limb assistive devices using ultrasound sensing, which may eventually enhance the intuitiveness and functionality of these devices towards a "free form" control approach.

## I. INTRODUCTION

Mobility problems affect the quality of life of an estimated 877 million people in the world [1]. For instance, individuals with a lower-limb amputation need to spend 30-60% higher metabolic energy compared to able-bodied individuals walking at the same speed, depending on the level of amputation [2]. Lower-limb assistive devices have the potential to enhance function of these individuals during activities of daily living; yet, up to 40% of the users of lower-limb prostheses report difficulty in controlling their prosthesis [3], [4]. Hence, an intuitive control behavior is still required to achieve seamless integration of human and assistive devices.

Lower-limb assistive devices are often controlled using finite-state machines or pattern-recognition algorithms [4], where there are several controllers responsible for different segments of a gait cycle and tasks, and pattern-recognition is used to switch between the underlying controllers [4], [5]. While this control scheme has enabled the use of assistive devices for ambulation on various surfaces [6], it is inherently different from the natural control of human ambulation by the nervous system. Natural limb movements are fluid and continuously adapting to individual intent. Therefore, biologically inspired methods are required to facilitate intuitive volitional control of lower-limb assistive devices that can continuously adapt to the user's intent.

Intuitive human-machine interfaces for direct volitional control of assistive devices have the potential to address these limitations by continuously responding to the neuromuscular signals of the user [6], resembling a more natural behavior. For instance, direct myoelectric control has been investigated for continuous control during level walking [7], stair ambulation [8], non-weight-bearing knee flexion-extension [9], and postural control [10]. Transtibial amputees have further shown the ability to adapt their locomotor function to alter the mechanics of a prosthetic ankle under continuous myoelectric control [11], [12]. Therefore, intuitive interfaces that respond to neural input by the user provide the possibility for an approach that can be generalized across various tasks, including steady-state and transient ambulation as well as unstructured movements.

Ultrasound sensing has recently emerged as an intuitive interface that can access human neuromuscular information by measuring muscle activation and contraction [13], [14]. Kinetic and kinematic ultrasound features of muscle have been shown to correlate to muscle function and joint motion [13], [15]; hence, they may be used for continuous estimation of cyclic and volitional movements. Rabe *et al*. successfully demonstrated the potential of transverse ultrasound images of the quadriceps for continuous estimation of knee angular velocity, as well as hip, knee, and ankle moments during steady-state treadmill ambulation [16], [17]. Longitudinal ultrasound features of muscle have been used for continuous estimation of human ankle moment [18], [19]; and to estimate and predict the knee kinematics, both during non-weight-bearing volitional movements [20], [21]. Based on the promise shown by ultrasound sensing as

This work was supported by the NSF grant 1925343 and the NIH grant R01EB025841.
M.H. Jahanandish and A. Srinivas are with the Bioengineering and Mechanical Engineering Departments, respectively, University of Texas at Dallas, Richardson, TX, USA. K. G. Rabe is with the Department of Biomedical Engineering, The University of Texas at Austin, Austin, TX, USA. N. P. Fey is with the Biomedical and Mechanical Engineering Departments, The University of Texas at Austin, Austin, TX, USA
K. Hoyt is with the Department of Bioengineering, University of Texas at Dallas, Richardson, TX, USA (e-mail: kenneth.hoyt@utdallas.edu).

an intuitive interface for continuous estimation of motion, more investigation towards an approach that can eventually generalize across various tasks is desirable.

Here, we present a novel method to encode spatiotemporal longitudinal ultrasound features of the rectus femoris (RF) and vastus intermedius (VI) muscles within a task-invariant learning paradigm for continuous estimation of steady-state and transient ambulation kinematics. We *hypothesize* that spatiotemporal ultrasound features of the proximal muscles can be used for task-invariant learning of the knee joint kinematics with an accuracy that is comparable to a task-specific learning paradigm.

## II. MATERIALS AND METHODS

### A. Subjects and Experiment

Seven healthy able-bodied individuals participated in the present study (4 males, 3 females; mean (SD) age: 29.6 (12.2) years). Dynamic anatomical data of the subjects, including joint movements, were recorded using a Vicon motion capture system that tracked the spatial location of the reflective markers located on the lower limbs, pelvis, and trunk of the individuals (spatial accuracy: ±0.142 mm). A custom 3D printed ultrasound probe holder was placed longitudinally over the RF muscle of the non-dominant leg, Fig. 1(a). Ultrasound images of the RF and VI muscles were captured using a handheld wearable ultrasound device (Lonshine Technologies, China), Fig. 1(b). Standard B-mode ultrasound images were collected with a sampling rate of 20 Hz in real-time using a transmit frequency of 7.5 MHz and a dynamic range of 50 dB. A custom software interface was used to stream the frames to a computer in real-time and timestamp each frame with a resolution of 1 ms.

The participants walked on a Bertec instrumented treadmill (Bertec, OH, USA), simulating level, and 10º incline or decline walking. Subjects completed one-minute trial for each task. Each subject chose a comfortable walking speed before the start of data collection; therefore, treadmill speed varied for different subjects. The mean (SD) speed of the treadmill during the three different tasks was: 1) level-walk: 0.76 (0.09) m/s, 2) incline: 0.59 (0.07) m/s, and 3) decline: 0.56 (0.06) m/s.

Five stair ambulation trials were performed on a 4-step staircase where the subjects were instructed to start with over ground walking, take one stride of level walking, and then transition to stair ambulation with the leading leg of their choice. Subjects were asked to perform two repetitions of a 4-stair ascent followed by a 4-stair descent with the reciprocal gait during each trial, for a total of ten 4-stair ascents and ten 4-stair descents per subject. The heel-strike and toe-off events were identified within Visual3D software (C-Motion, MD, USA) and the gait events specific to the stair transition strides were marked for further analysis. The mean (SD) number of strides during each ambulation task are as follows, level-walk: 40.7 (4.1), incline: 37.7 (6.4), decline: 41.7 (3.1), steady-state stair ascent: 12.6 (1.8), transient stair ascent: 7.4 (1.8), steady-state stair descent: 14.8 (2.2), and transient stair descent 5.2 (2.2).

### B. Spatiotemporal Ultrasound Features

Ultrasound echogenicity has been shown to correlate with muscle contraction and joint motion [15]. Therefore, ultrasound images can be used to create time-intensity features of the muscle which could be used as predictors of the motion [22]. The thickness of muscle aponeuroses has been determined to vary between 1-3 mm [23]; hence, the mean image intensity of 3 mm x 3 mm kernels was extracted from each ultrasound frame to create a feature set consisting of $n$ values, where $n$ is the number of kernels per frame. The feature set was further reshaped into a $n$x1 column representing time-intensity features of the superficial muscle tissue at the top and the features of deep muscle tissue at the bottom. To grasp the contribution of temporal information from the sequence of ultrasound images during a gait cycle, the time-derivative of the intensity features was calculated between every consecutive frame pairs. Hereafter, we refer to these time-derivatives as temporal features. Figures 2(a) and 2(b) illustrate the progression of the intensity and temporal features across the gait cycle of a representative subject, respectively, where each pixel column represents the ultrasound features at a certain time point.

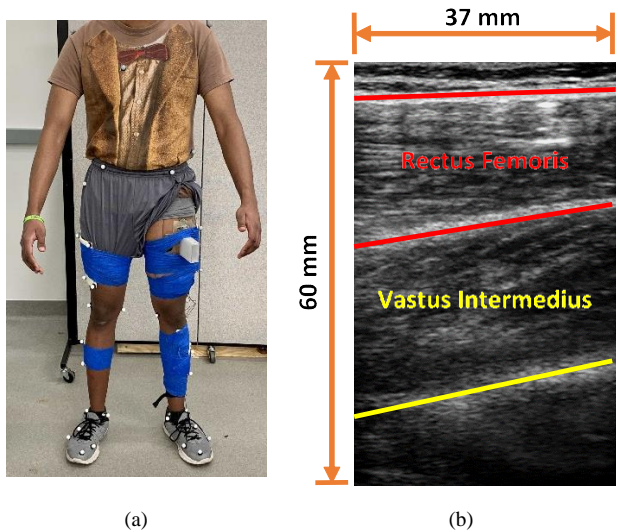

Fig. 1. (a) Experimental setup on a human subject. 3D printed probe holder was placed over the rectus femoris (RF) muscle of the non-dominant leg. (b) Sample longitudinal ultrasound image highlighting the RF and vastus intermedius (VI) muscles.

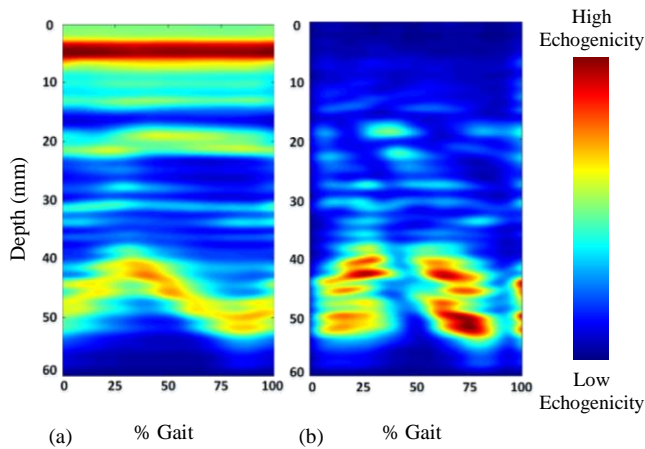

Fig. 2. The progression of the (a) intensity and (b) temporal ultrasound features of the RF and VI muscles during the level walking gait of a sample subject. The features are averaged across 1 min of walking strides.

## C. Motion Estimation: Task-Specific vs Task-Invariant

The ultrasound feature sets were used to investigate the feasibility of continuous task-invariant learning of knee kinematics during the steady-state and transient ambulation trials. Gaussian Process Regression (GPR) models with a quadratic kernel were trained using the ultrasound intensity and temporal features as predictors to continuously estimate the kinematics of the knee joint as the target variables based on two different task-specific and task-invariant learning paradigms. Task-specific learning was achieved by training a GPR model for each specific task. Conversely, a single GPR model was trained using all ambulation tasks during the task-invariant learning. Additionally, all models were trained with and without temporal features. The trained models were then evaluated using a leave-N-strides-out cross-validation, where 5-6 consecutive strides of the treadmill trials were left out for testing during each round of cross-validation (~20% of all the strides). In the case of stair ambulation, a leave-one-trial-out scheme was used during each round of cross-validation (20% of all the repetitions). Both task-specific and task-invariant models were tested separately on the steady-state stair strides as well as the walk-to-stair and stair-to-walk transition strides.

The overall effect of the learning paradigm and the effect of temporal features on the accuracy of estimation were statistically evaluated using the repeated-measures two-way analysis of variance (ANOVA) test. Multiple *posthoc* comparisons were performed to statistically compare each condition, and the *p*-values were adjusted using a Bonferroni correction for multiple comparisons ($\alpha = 0.05$).

## III. RESULTS

### A. Learning the Continuous Knee Kinematics during Ambulation

The root mean square error (RMSE) of continuous task-specific and task-invariant learning of the knee joint kinematics was evaluated for level walking, incline walking, decline walking, stair ascent and stair descent. For stair ascent and stair descent trials, both steady-state and transition strides were included for estimation of overall performance. Overall, the average RMSEs of knee angle and angular velocity estimation across tasks and subjects were 7.06 (1.29)° and 53.1 (8.1)°/sec for the task-invariant learning paradigm, respectively. These results were not significantly different from the average RMSEs of 6.00 (1.62)° and 51.8 (9.2)°/sec for the task-specific models ($p = 0.11$ for angle and $p = 0.66$ for angular velocity estimations).

TABLE I. MEAN (SD) RSME (DEG) OF KNEE ANGLE ESTIMATION ACROSS AMBULATION TASKS AND LEARNING PARADIGMS

| Task: | Level | | Incline | | Decline | | Stair Ascent | | Stair Descent | |
|---|---|---|---|---|---|---|---|---|---|---|
| Feature Set: | Intensity | Temporal [d] | Intensity | Temporal [d] | Intensity | Temporal [c] | Intensity | Temporal [d] | Intensity | Temporal [c] |
| Task-Specific | 4.9 (0.5) | 4.0 (0.9) | 3.8 (0.7) | 2.7 (0.6) | 4.6 (1.0) | 3.8 (0.5) | 11.5 (2.3) | 9.8 (2.2) | 12.4 (2.5) | 11.6 (2.8) |
| Task-Invariant | 5.2 (1.1) | 4.7 (1.0) | 4.0 (1.1) | 3.5 (0.7) [b] | 5.4 (1.1) [b] | 4.4 (0.7) | 12.5 (3.1) | 10.7 (2.7) | 13.7 (2.9) [a] | 12.1 (3.0) |

[a] and [b] indicate $p < 0.05$ and $p < 0.01$ for the significant effect of the learning paradigm on the RMSE of estimations, respectively. [c] and [d] indicate $p < 0.05$ and $p < 0.01$ for the main significant effect associated with incorporating the temporal features for training, respectively.

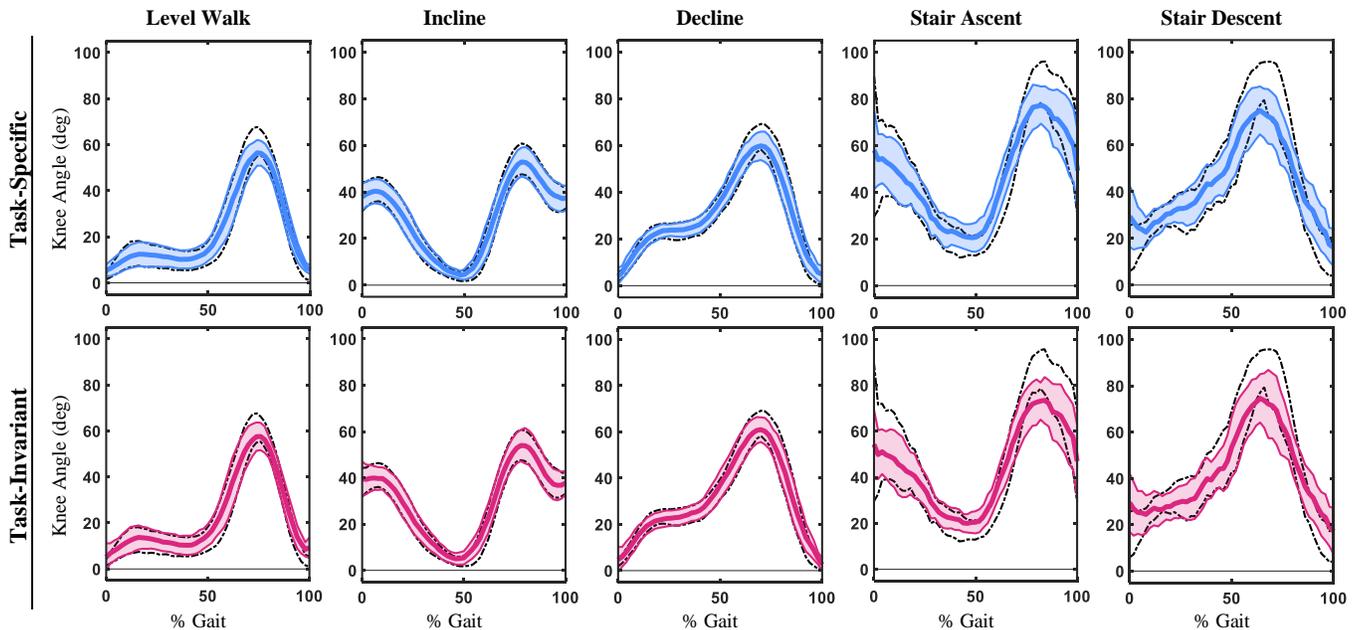

Fig. 3. Continuous estimation of the knee joint angle based on the task-specific (first row) and task-invariant (second row) learning paradigms. The dashed black lines show the mean ± 1 SD recorded trajectories and the solid colored lines show the predicted trajectories. The trajectories are averaged across all subjects and the shaded areas represent mean ± 1 SD.

The first rows in Tables I and II show the RMSE of the continuous task-specific estimation of the knee angle and angular velocity, respectively. The second rows in Tables I and II represent continuous task-invariant learning of the ambulation kinematics. Additionally, the RMSEs of the models trained with and without the temporal ultrasound feature sets are shown side-by-side for each task. Statistical analysis revealed that continuous task-invariant learning of ambulation kinematics did not introduce a significant difference in the RMSE of estimations ($p = 0.12$ for angle estimation, and $p = 0.66$ for angular velocity estimation). It was further discovered that including the temporal ultrasound features of the muscle not only significantly improved reduced the RMSE ($p < 0.05$) in comparison to intensity features only, but also helped reduce the difference between the RMSE of task-specific and task-invariant learning models. Similarly, the first rows in Figs. 3 and 4 illustrate the continuous estimation of the knee joint angle and angular velocity based on task-specific learning models. Second rows in Figs. 3 and 4 show the continuous estimation of ambulation kinematics using task-invariant learning models. While there are a few points during the gait cycle where the task-invariant learning does not perform as well as the task-specific learning (e.g. at the very beginning of the stair descent gait, Fig. 3), the estimations have similar overall trends for all tasks when comparing the two learning paradigms. In the case of stair ambulation, both steady-state and transition strides were included for estimation of overall performance; therefore, the standard deviation bands are larger due to the higher stride-to-stride variability.

### B. The Effect of Transient Stair Ambulation on Task-Invariant Learning

Tables III and IV report the RMSEs of the knee angle and angular velocity estimations when evaluating transition and steady-state strides separately, using each of the two learning paradigms and ultrasound feature sets. Figure 5 illustrates the continuous estimation of knee kinematics during the stair ascent task averaged across all subjects. Data from the stair descent task have not been shown for brevity. The strides shown in Fig. 5 are time-normalized based on the gait progression, where the walk-to-stair, steady-state, and stair-to-walk strides are shown by [-100%-0%], [0%-100%], and [100%-200%], respectively. Interestingly, the combination of using task-invariant learning models along with the temporal ultrasound features consistently improved the estimation accuracy of both joint angle and angular velocity during transition strides, compared to the task-specific learning paradigm (Tables III and IV). However, this effect was not significant ($p = 0.68$).

TABLE II.  MEAN (SD) RSME (DEG/S) OF KNEE ANGULAR VELOCITY ESTIMATION ACROSS AMBULATION TASKS AND LEARNING PARADIGMS

| Task: | Level | | Incline | | Decline | | Stair Ascent | | Stair Descent | |
|---|---|---|---|---|---|---|---|---|---|---|
| **Feature Set:** | Intensity | Temporal | Intensity | Temporal | Intensity | Temporal | Intensity | Temporal [d] | Intensity | Temporal |
| **Task-Specific** | 40.9 (7.5) | 36.8 (7.5) | 29.8 (11.5) | 23.8 (5.4) | 37.4 (7.9) | 36.1 (6.4) | 71.1 (16.5) | 65.8 (16.2) | 101.8 (22.2) | 96.3 (20.7) |
| **Task-Invariant** | 41.3 (6.8) | 42.8 (6.3) | 25.8 (3.2) | 26.7 (4.8) | 42.3 (4.7) | 39.8 (4.5) | 72.8 (15.6) | 63.1 (14.2) | 99.6 (18.9) | 92.9 (19.2) |

[d] indicates $p < 0.01$ for the main significant effect associated with incorporating the temporal features for training.

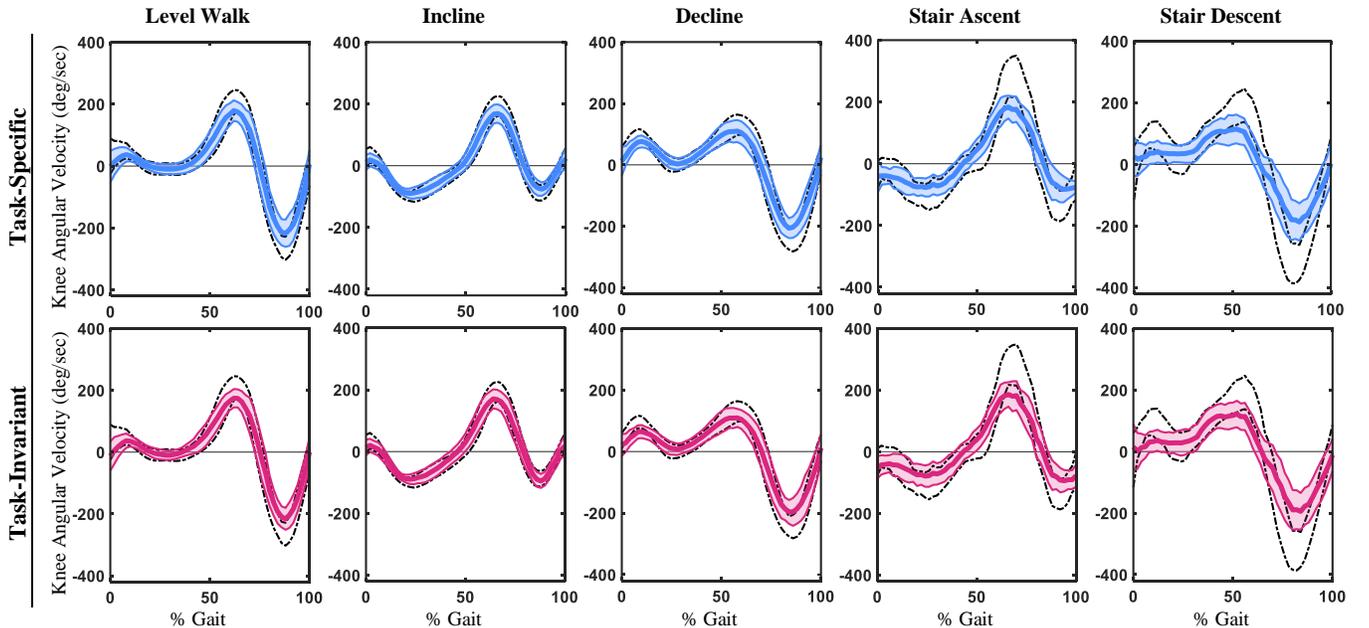

Fig. 4. Continuous estimation of the knee joint angular velocity based on the task-specific (first row) and task-invariant (second row) learning paradigms. The dashed black lines show the mean ± 1 SD recorded trajectories and the solid colored lines show the predicted trajectories. The trajectories are averaged across all subjects and the shaded areas represent mean ± 1 SD.

## IV. Discussion

In this study, we demonstrated the feasibility of using the spatiotemporal ultrasound features of muscle for the continuous task-invariant learning of knee joint kinematics during steady-state and transient ambulation.

The task-invariant learning paradigm demonstrated statistically comparable performance to task-specific models during various ambulation tasks, especially after the temporal ultrasound features of the muscles were included for training. However, it appears that letting the models learn from the steady-state treadmill strides specifically improves the continuous estimation of knee kinematics during transient stair ambulation (Tables III and VI). This effect was noticeable in the continuous knee angle estimation during the first 50% of the walk-to-stair strides, where task-invariant learning presented a more accurate estimation of the knee angle compared to task-specific models (see Fig. 5).

Previous work on the volitional myoelectric control of a knee prosthetic has shown RMSE of 6.2° for trajectory tracking during non-weight-bearing flexion/extension movements, compared to a RMSE of 5.2° for the intact knee [9]. We have also shown in a previous work that an ultrasound-based approach can continuously estimate the knee joint kinematics with an RMSE of 7.45° during the same movement [20]. Recently, Rai *et al.* proposed a mode-free control method that can estimate the ankle joint angle with an RMSE of 7° during level-walking and stair ambulation [24]. However, their approach does not integrate the neuromuscular signals from the user and requires instrumentation of both lower limbs with inertial sensors. Our proposed ultrasound-based approach improves upon these works by addressing their limitations while maintaining a mean RMSE of 7.06° for the task-invariant learning of steady-state and transient ambulation kinematics.

While the task-invariant models fully capture the motion trajectories regardless of the activity, they seem to fail to capture the full range of motion during the knee swing flexion of the steady-state stair ambulation (Figs. 3 and 5). However, the task-invariant models still performed better than the performance metrics threshold for smooth stair ambulation by Azocar *et al.* [25]. They have shown that knee-swing flexion angles of 71.9° and 70.5° are effective for the reciprocal gait during stair ascent and descent, respectively. Therefore, our ultrasound-based task-invariant models with average estimated knee-swing flexions of 81.9° for stair ascent and 79.0° for stair descent would still maintain smooth reciprocal gait during stair ambulation. Interestingly, the task-invariant models did not have any trouble learning the full range of motion during the transition strides.

TABLE III. Mean (SD) RSME (deg) of Knee Angle Estimation During Steady-State and Transient Stair Ambulation

| Task: | Stair Ascent | | | | | | Stair Descent | | | | | |
|---|---|---|---|---|---|---|---|---|---|---|---|---|
| | Walk-to-Stair | | Steady-State | | Stair-to-Walk | | Walk-to-Stair | | Steady-State | | Stair-to-Walk | |
| Feature Set: | Intensity | Temporal [c] | Intensity | Temporal [d] | Intensity | Temporal [d] | Intensity | Temporal [d] | Intensity | Temporal | Intensity | Temporal |
| Task-Specific | 18.0 (5.2) | 14.5 (3.5) | 15.7 (1.6) | 13.0 (2.0) | 14.5 (3.3) | 13.6 (3.6) | 18.4 (6.5) | 17.2 (8.2) | 12.7 (2.7) | 11.6 (2.1) | 15.5 (4.6) | 15.8 (4.9) |
| Task-Invariant | 17.2 (5.1) | 14.4 (3.4) | 16.7 (1.8) | 14.2 (2.3) | 15.3 (3.2) | 13.1 (3.2) | 18.3 (5.8) | 16.6 (6.6) | 15.3 (3.5) [b] | 12.9 (2.3) [a] | 17.4 (5.0) | 15.9 (4.8) |

[a] and [b] indicate $p < 0.05$ and $p < 0.01$ for the significant effect of the learning paradigm on the RMSE of estimations, respectively. [c] and [d] indicate $p < 0.05$ and $p < 0.01$ for the main significant effect associated with incorporating the temporal features for training, respectively.

TABLE IV. Mean (SD) RSME (deg/s) of Knee Angular Velocity Estimation During Steady-State and Transient Stair Ambulation

| Task: | Stair Ascent | | | | | | Stair Descent | | | | | |
|---|---|---|---|---|---|---|---|---|---|---|---|---|
| | Walk-to-Stair | | Steady-State | | Stair-to-Walk | | Walk-to-Stair | | Steady-State | | Stair-to-Walk | |
| Feature Set: | Intensity | Temporal | Intensity | Temporal [c] | Intensity | Temporal [d] | Intensity | Temporal [d] | Intensity | Temporal [c] | Intensity | Temporal |
| Task-Specific | 80.2 (21.5) | 74.8 (22.6) | 77.4 (21.4) | 73.8 (17.9) | 113.3 (28.7) | 99.2 (31.1) | 128.4 (28.2) | 113.1 (23.4) | 101.5 (26.1) | 95.5 (21.8) | 114.3 (26.3) | 109.4 (23.1) |
| Task-Invariant | 79.4 (19.9) | 74.4 (18.6) | 78.3 (19.3) | 70.9 (16.0) | 115.9 (24.3) | 95.2 (30.1) [a] | 118.5 (25.0) [a] | 106.8 (30.2) | 103.9 (23.9) | 94.6 (20.5) | 108.2 (19.5) | 108.9 (18.4) |

[a] indicates $p < 0.05$ for the significant effect of the learning paradigm on the RMSE of estimations. [c] and [d] indicate $p < 0.05$ and $p < 0.01$ for the main significant effect associated with incorporating the temporal features for training, respectively.

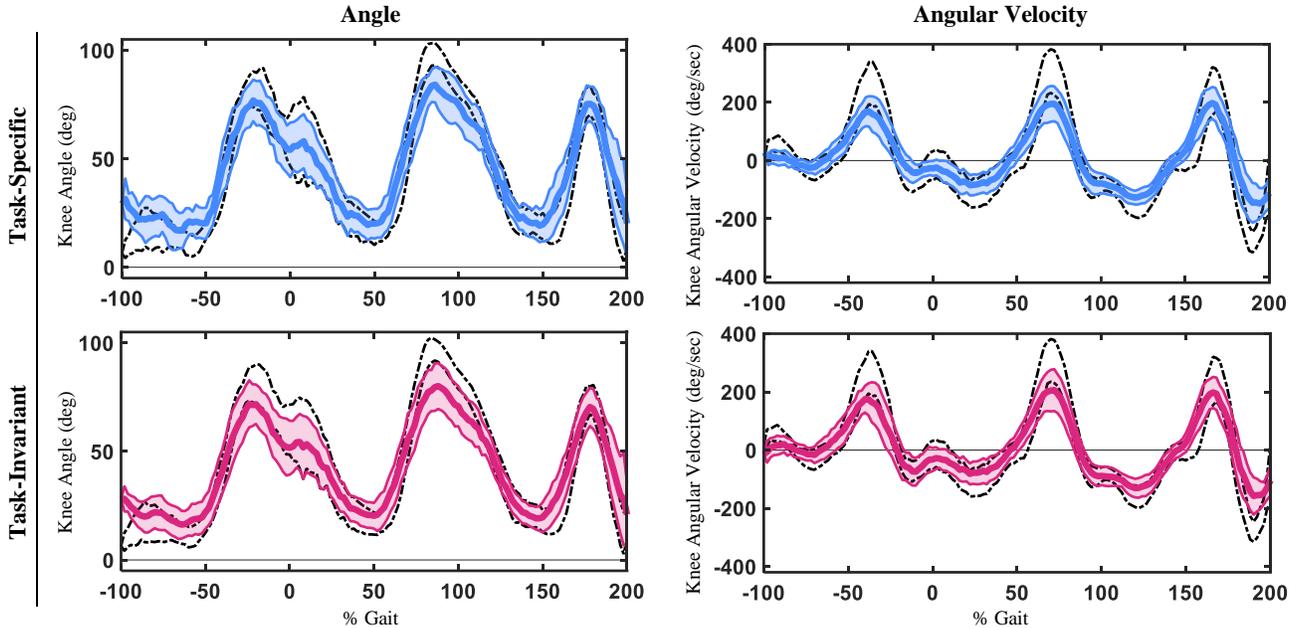

Fig. 5. Continuous estimation of the knee joint angle (left column) and angular velocity (right column) during transient stair ambulation based on the task-specific (first row) and task-invariant (second row) learning paradigms. The strides are time-normalized based on the gait progression, where the walk-to-stair, steady-state, and stair-to-walk strides are shown by [-100%-0%], [0%-100%], and [100%-200%], respectively. The dashed black lines show the mean ± 1 SD recorded trajectories and the solid colored lines show the predicted trajectories. The trajectories are averaged across all subjects and the shaded areas represent mean ± 1 SD.

*The Effect of Temporal Ultrasound Features*

While the task-invariant models trained using only the ultrasound intensity features showed statistically comparable performance to most task-specific models, including the temporal ultrasound features within the task-invariant paradigm resulted in comparable performance for all ambulation tasks. Moreover, incorporating the temporal ultrasound features consistently narrowed the margin of error between continuous estimation of the two learning paradigms, and helped task-invariant models to outperform their counterparts in the continuous estimation of transient ambulation. This is consistent with previous studies that have shown the positive effect of incorporating the time history of neural and mechanical signals on the accuracy of motion control and estimation during ambulation [26]–[28].

*Limitations and Future Work*

This study demonstrates the feasibility of an ultrasound-based approach for the continuous task-invariant learning of steady-state and transient ambulation using able-bodied subjects. Future work needs to focus on translating the same paradigm to the target populations for lower-limb assistive devices, i.e. populations with mobility-related pathology or limb-loss. While our approach demonstrated useful for continuous transient ambulation, there are still some common volitional activities such as sit-to-stand, non-weight-bearing, and unstructured movements that need to be incorporated within a task-invariant learning paradigm for a truly "free form" control scheme. The positive effect of incorporating the temporal features on the accuracy of task-invariant learning suggests that the sequence-to-sequence prediction models might prove particularly useful to encode the spatiotemporal features of ultrasound data for the task-invariant learning of continuous ambulation. Furthermore, subject-specific models were trained for this study. While it has been shown difficult to rely solely on neuromuscular signals [17], [29] for a subject-independent approach, it may be feasible to combine the temporal ultrasound features with anatomically normalized intensity features to obtain the desired performance based on a subject-independent approach. Due to the high between-subject variability of the neuromuscular signals, partially subject-independent schemes with a larger sample size might be more feasible for task-invariant learning of continuous volitional ambulation.

## V. CONCLUSION

This study demonstrated the feasibility of using spatiotemporal ultrasound features of the RF and VI muscles for task-invariant learning of continuous knee kinematics during various steady-state and transient ambulation tasks. It was observed that the continuous joint kinematics can be learned using a task-invariant paradigm with statistically comparable accuracy to a task-specific paradigm. Statistical analysis further revealed that incorporating the temporal ultrasound features significantly improves the accuracy of continuous estimations ($p < 0.05$). These results motivate the future work toward using ultrasound sensing as an intuitive interface for a more biologically inspired "free form" control approach, and its implementation on lower-limb assistive devices. Intuitive interfaces that can continuously respond to the user's intent may eventually enhance the clinical outcome and functionality of assistive devices.